\documentclass{article}

\usepackage{PRIMEarxiv}

\usepackage[utf8]{inputenc} 
\usepackage[T1]{fontenc}    
\usepackage{hyperref}       
\usepackage{url}            
\usepackage{booktabs}       
\usepackage{amsfonts}       
\usepackage{nicefrac}       
\usepackage{microtype}      
\usepackage{lipsum}
\usepackage{fancyhdr}       
\usepackage{graphicx}       
\usepackage{amsmath}
\usepackage{subcaption}
\usepackage{multirow}
\usepackage{natbib} 
\usepackage{float}
\usepackage{tabularx}
\graphicspath{{media/}}     

\title{Trajectory Geometry of Transformer Representations Across Layers}

\author{
  Vishal Pandey\thanks{Code available at: \url{https://github.com/Vishal-sys-code/latent-trajectories}}\\
  London, UK\\
  \texttt{vishal@metriqual.com} \\
  \And
  Gopal Singh\\
  Athens, GR\\
  \texttt{gopal@metriqual.com} \\
  \And
  Yacine Mahdid \\
  Montreal, CA\\
  \texttt{yacine@datadom.co} \\
}

\begin{document}
\maketitle

\begin{abstract}
Understanding how transformer representations evolve across layers not merely what they encode remains an open problem in mechanistic interpretability. We recast the transformer forward pass as a discrete population trajectory through a high-dimensional representation manifold, drawing on geometric tools from computational neuroscience. Rather than probing for pre-specified features, we characterize the intrinsic geometry of these trajectories using five metrics computed directly in the ambient space: trajectory length, curvature, a semantic convergence index, layerwise cosine similarity, and representational stability. Across three model families (GPT-2, TinyLlama, Qwen2.5) and five semantically controlled prompt families, we report four principal findings. First, semantically related prompts undergo statistically significant trajectory convergence in middle-to-late layers, with peak convergence indices of 0.41--0.58 across architectures ($p < 0.001$, Mann-Whitney U), consistent with attractor-like dynamics. Second, reasoning and analogy tasks produce trajectories of significantly greater curvature than lexical variation tasks (0.71--0.83\,rad vs.\ 0.27--0.31\,rad), suggesting that mean curvature encodes computational complexity. Third, ambiguous tokens exhibit measurable trajectory bifurcation, a clean disambiguation signature with up to $5.6\times$ representational separation by the final layer, absent in unambiguous controls. Fourth, layerwise cosine similarity reveals a universal three-phase computational structure: encoding, elaboration, and output preparation, whose boundaries are consistent across all three architectures and align with layer ranges implicated in induction head formation and MLP based knowledge retrieval in prior mechanistic work. All four effects vanish under shuffled-layer and random-embedding controls, confirming they are intrinsic to learned computation. We release a fully open-source, model-agnostic pipeline and argue that trajectory geometry constitutes a principled, probe-free lens for mechanistic interpretability.
\end{abstract}

\keywords{representation geometry \and transformer interpretability \and neural manifolds \and trajectory analysis \and population dynamics \and mechanistic interpretability}

\section{Introduction}
\label{sec:introduction}

While modern transformers \cite{vaswani2017attention} achieve remarkable 
performance across diverse natural language tasks, the internal mechanisms 
governing their representations remain largely opaque. Existing interpretability  approaches operate primarily in two regimes: \textit{mechanistic} analyses that  trace individual attention heads and circuit-level computations  \cite{olsson2022incontext, elhage2021mathematical}, and \textit{static} analyses  that probe layer-wise embeddings for pre-specified linguistic features  \cite{tenney2019bert, jawahar2019bert}. Both paradigms treat each layer as an independent snapshot, ignoring the continuous geometric structure of how representations \textit{evolve} across the depth of the network.

We propose a complementary perspective: the transformer forward pass as a 
\textbf{discrete population trajectory} through a high-dimensional representation  manifold. Rather than probing for what is encoded at a given layer, we ask how representations travel from input to output, characterizing the geometry of the path itself. This framing is inspired by population trajectory analyses in computational neuroscience \cite{vyas2020computation, cunningham2014dimensionality},  where collective neural activity is studied as an orbit on a low-dimensional manifold rather than as independent neuron-by-neuron measurements. We adapt this 
toolkit to artificial networks without claiming any correspondence to biological cognition.

Concretely, we define five metrics computed directly in the full ambient 
representation space, trajectory length, curvature, a semantic convergence index,  layerwise cosine similarity, and representational stability, and apply them to three transformer families (GPT-2 \cite{radford2019language}, TinyLlama \cite{zhang2024tinyllama}, Qwen2.5 \cite{qwen2025qwen25}) across five semantically controlled prompt families. We find that: (\textit{i}) semantically related representations converge toward attractor-like regions in middle-to-late layers; (\textit{ii}) reasoning tasks induce significantly higher trajectory curvature than surface-form variations; and (\textit{iii}) ambiguous tokens undergo measurable trajectory bifurcation at a consistent network depth. All effects survive four rigorous control experiments.

\paragraph{Contributions.}
\begin{itemize}
    \item A \textbf{trajectory-geometric framework} for transformer 
    interpretability, defining five probe-free, high-dimensional metrics that 
    characterize representation dynamics across layers.
    
    \item \textbf{Four empirical findings} semantic convergence into attractor basins, curvature as a probe-free complexity readout, disambiguation as trajectory bifurcation, and a universal three-phase computational structure, each replicated across three architectures and validated against shuffled-layer, random-embedding, and random-label controls.

    \item An \textbf{open-source pipeline} (\texttt{github.com/Vishal-sys-code/latent-trajectories}) enabling trajectory analysis for any causal language model without requiring probing classifiers or fine-tuning.
    
    \item A \textbf{theoretical bridge} connecting the dynamical systems 
    literature in computational neuroscience to mechanistic interpretability 
    in deep learning.
\end{itemize}

\section{Related Work}
\label{sec:related_work}
The present work sits at the intersection of mechanistic interpretability, representational geometry, computational neuroscience, and probing-based analyses of language models. We review each thread and position our contributions relative to their limitations.

\paragraph{Mechanistic Interpretability:} A substantial body of work seeks to reverse-engineer transformers by isolating discrete computational subcomponents. Elhage et al.\ \cite{elhage2021mathematical} established a mathematical framework for transformer circuits, enabling the discovery of induction heads \cite{olsson2022incontext} attention patterns responsible for in-context learning. Subsequent work localized factual recall to specific MLP layers \cite{meng2022locating, geva2021transformer} and identified superposition as a mechanism by which a single neuron encodes multiple features \cite{elhage2022superposition}. Nanda et al.\ \cite{nanda2023progress} demonstrated that mechanistic analysis can track the emergence of algorithmic structure during training. While these approaches successfully identify \textit{which} components execute specific computations, they analyze components in isolation and do not characterize the \textit{global geometric consequence} of all components acting jointly across the full layer sequence. Our work is explicitly complementary: 
where circuit analysis asks \textit{who}, trajectory geometry asks \textit{how the whole system moves}.

\paragraph{Probing Classifiers and Layer-wise Analysis:} Probing methods train lightweight classifiers on frozen representations  to test whether a pre-specified feature is linearly decodable at a given layer \cite{alain2016understanding}. Applied to BERT, Tenney et al.\  \cite{tenney2019bert} and Jawahar et al.\ \cite{jawahar2019bert} showed that syntactic structure is resolved in early layers while semantic content emerges in later ones. Geva et al.\ \cite{geva2022transformer} demonstrated that MLP layers promote output vocabulary concepts in later layers via a key-value retrieval mechanism. The logit lens \cite{nostalgebraist2020logit} extends this by projecting intermediate representations directly into vocabulary space to track early prediction formation. A fundamental limitation of all probing approaches is that they require the analyst to specify \textit{in advance} what to look for. Our trajectory framework requires no such specification: we characterize the geometry of the path itself, making it discovery-oriented rather than confirmatory.

\paragraph{Representational Geometry and Similarity:} Centered Kernel Alignment (CKA) \cite{kornblith2019similarity} provides a principled measure of representational similarity between layers and models, revealing that representations converge in deeper layers across architectures. The Platonic Representation Hypothesis \cite{huh2024platonic} extends this, arguing that models trained on different modalities and objectives converge toward a shared statistical model of reality. Representational Similarity Analysis (RSA), originating in systems neuroscience \cite{kriegeskorte2008representational}, compares geometry across conditions via dissimilarity matrices and has been applied to compare biological and artificial representations. Elhage et al.\ \cite{elhage2022superposition} demonstrated that features can be geometrically superposed, with representations occupying non-orthogonal directions to exceed dimensional capacity. These works establish that transformer representations have rich geometric structure, but analyze it at fixed layers. We extend this program by treating the \textit{layer sequence itself} as a geometric object, a trajectory, whose shape encodes computational meaning.

\paragraph{Neural Population Dynamics and Manifold Theory:} In computational neuroscience, neural population activity is studied as trajectories on low-dimensional manifolds embedded in high-dimensional firing-rate space \cite{cunningham2014dimensionality, vyas2020computation}. Key findings include: motor cortex trajectories exhibit consistent geometric structure during movement preparation \cite{shenoy2013cortical}; trajectory curvature covaries with task complexity during flexible decision-making \cite{remington2018flexible}; and attractor dynamics describe how populations converge to stable states encoding decisions or memories \cite{hopfield1982neural}. Gallego et al.\ \cite{gallego2017neural} showed that motor cortex activity is confined to a low-dimensional neural manifold largely invariant to task conditions. While researchers have drawn loose analogies between these dynamics and transformer computation \cite{sussillo2014neural}, no prior work has operationalized trajectory length, curvature, and convergence indices as quantitative metrics applied systematically to transformer hidden states with rigorous statistical controls.

\paragraph{Dynamical Systems Views of Deep Networks:} Several works have analyzed deep networks through a dynamical systems lens. Raghu et al.\ \cite{raghu2017svcca} used SVCCA to show that representations in deep networks stabilize from the bottom up during training. Morcos et al.\ \cite{morcos2018insights} demonstrated that networks with more similar representations generalize better. Recent work on neural collapse \cite{papyan2020prevalence} shows that last-layer representations collapse to class means at convergence, a specific form of attractor dynamics in the final layer. Our work generalizes this picture: rather than studying collapse at a single layer or similarity between fixed checkpoints, we track the full  geometric evolution of representations \textit{during the forward pass}, 
revealing phase-transition structure that is invisible to any single-layer analysis.

\paragraph{Collective Gap:} Taken together, prior work has established that transformer representations are geometrically structured (CKA, RSA, superposition), that specific components perform specific computations (circuits, probing), and that deep networks exhibit dynamical phenomena (neural collapse, SVCCA stabilization). What is missing is a unified, probe-free, trajectory-level characterization of how representations evolve \textit{continuously} across all layers, one that connects individual geometric properties (length, curvature, convergence) to specific computational behaviors (semantic clustering, reasoning complexity, disambiguation). This paper fills that gap.

\section{Methodology}
\label{sec:methodology}

\begin{figure}[htbp]
    \centering
    \includegraphics[width=0.4\linewidth]{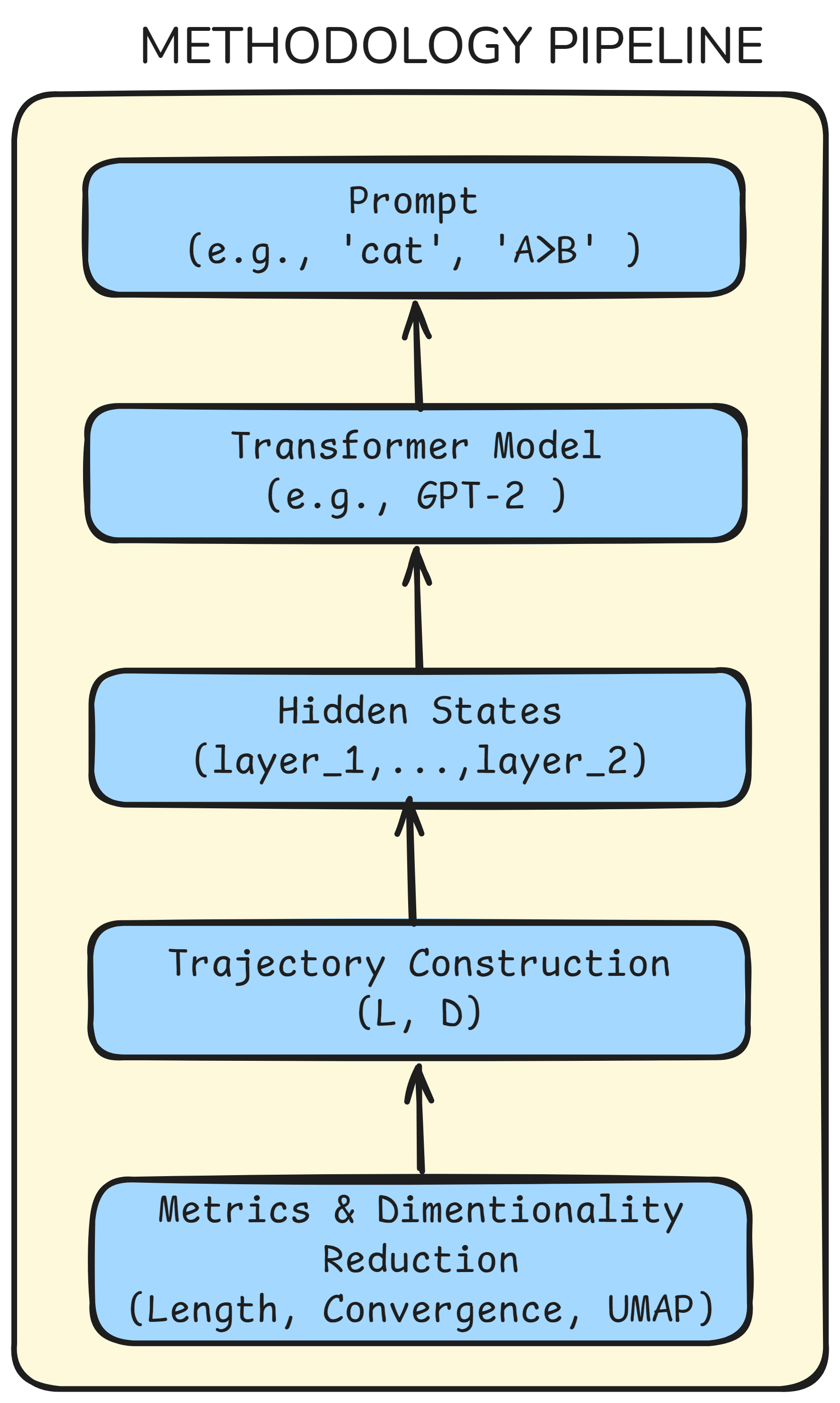}
    \caption{Analytical pipeline. From prompt input to hidden state 
    extraction, high-dimensional metric computation, statistical 
    validation, and visualization.}
    \label{fig:pipeline}
\end{figure}

Our analytical framework consists of three sequential stages: (1) hidden state extraction, (2) high-dimensional geometric metric computation, and (3) statistical validation. All geometric metrics are computed directly in the full ambient representation space $\mathbb{R}^d$; dimensionality reduction is applied strictly for visualization and plays no role in any reported result. Figure~\ref{fig:pipeline} illustrates the complete pipeline.

\subsection{Trajectory Extraction}
\label{subsec:extraction}

Given a transformer language model $f_\theta$ with $L$ layers and hidden dimension $d$, let $H^{(l)} \in \mathbb{R}^{n \times d}$ denote the matrix of hidden states at layer $l$ for an input sequence of $n$ tokens. We define the \textbf{trajectory representation} of an input at layer $l$ as the mean-pool over non-padding token positions:
\begin{equation}
    \mathbf{h}^{(l)} = \frac{1}{n} \sum_{i=1}^{n} H^{(l)}_i 
    \in \mathbb{R}^d
\end{equation}
Mean pooling is chosen over last-token extraction to produce sequence-level representations that are robust to positional artifacts and consistent across prompts of varying length. The \textbf{trajectory} of a prompt is the ordered sequence of these representations across all layers:
\begin{equation}
    \tau = \left(\mathbf{h}^{(0)},\, \mathbf{h}^{(1)},\, \dots,\, 
    \mathbf{h}^{(L)}\right) \in \left(\mathbb{R}^d\right)^{L+1}
\end{equation}
Layer 0 corresponds to the input embedding prior to any transformer block computation. We retain it as the trajectory origin to capture the full representational transformation from raw token embeddings to contextualized outputs. All hidden states are extracted using \texttt{output\_hidden\_states=True} and stored as \texttt{.pt} tensors indexed by prompt ID for full reproducibility.

\subsection{Geometric Metrics}
\label{subsec:metrics}

We define five metrics that characterize distinct geometric properties of trajectories, all computed in the full $\mathbb{R}^d$ ambient space.

\paragraph{Trajectory Length:} The total Euclidean displacement accumulated across layers:
\begin{equation}
    \mathcal{L}(\tau) = \sum_{l=0}^{L-1} 
    \left\lVert \mathbf{h}^{(l+1)} - \mathbf{h}^{(l)} \right\rVert_2
\end{equation}
Large $\mathcal{L}$ indicates substantial representational transformation; near-zero increments at a layer suggest an approximately identity computation at that depth.

\paragraph{Trajectory Curvature:} The local curvature at layer $l$ is defined as the turning angle between consecutive displacement vectors:
\begin{equation}
    \kappa^{(l)} = \arccos\!\left(
        \frac{\mathbf{v}^{(l)} \cdot \mathbf{v}^{(l+1)}}
             {\left\lVert\mathbf{v}^{(l)}\right\rVert_2 \,
              \left\lVert\mathbf{v}^{(l+1)}\right\rVert_2}
    \right), \quad
    \mathbf{v}^{(l)} = \mathbf{h}^{(l)} - \mathbf{h}^{(l-1)}
\end{equation}
Mean curvature over the trajectory is $\bar{\kappa}(\tau) = 
\frac{1}{L-1}\sum_{l=1}^{L-1} \kappa^{(l)}$. High curvature indicates non-linear traversal of representation space; low curvature indicates near-geodesic (straight-line) evolution. We hypothesize that curvature encodes computational complexity, with reasoning tasks producing significantly higher $\bar{\kappa}$ than surface-form variations.

\paragraph{Semantic Convergence Index:} For a semantic category $\mathcal{C} = \{\tau_1, \dots, \tau_k\}$, the convergence index at layer $l$ is:
\begin{equation}
    \text{CI}(l) = D_{\text{between}}(l) - D_{\text{within}}(l)
\end{equation}
where $D_{\text{within}}(l)$ is the mean pairwise Euclidean distance among representations in $\mathcal{C}$ at layer $l$, and $D_{\text{between}}(l)$ is the mean pairwise distance between $\mathcal{C}$ and all representations outside $\mathcal{C}$. Positive $\text{CI}(l)$ indicates semantic compression: members of the same category are more tightly clustered relative to inter-category distances, consistent with attractor-like convergence.

\paragraph{Layerwise Cosine Similarity:} The angular similarity between adjacent-layer representations:
\begin{equation}
    \text{SIM}(l) = 
    \frac{\mathbf{h}^{(l)} \cdot \mathbf{h}^{(l+1)}}
         {\left\lVert\mathbf{h}^{(l)}\right\rVert_2 \, 
          \left\lVert\mathbf{h}^{(l+1)}\right\rVert_2}
\end{equation}
Sharp drops in $\text{SIM}(l)$ identify layers undergoing significant directional change, computational phase transitions analogous to velocity discontinuities in neural population trajectories.

\paragraph{Representational Stability:} For a prompt $p$ and a lexical perturbation $p'$ (e.g., \textit{cat} $\to$ \textit{a cat}), stability at layer $l$ is defined as:
\begin{equation}
    \text{STAB}(l) = 
    \frac{\mathbf{h}^{(l)}_p \cdot \mathbf{h}^{(l)}_{p'}}
         {\left\lVert\mathbf{h}^{(l)}_p\right\rVert_2 \, 
          \left\lVert\mathbf{h}^{(l)}_{p'}\right\rVert_2}
\end{equation}
High stability indicates that surface, form variation is abstracted away by layer $l$; low stability indicates residual sensitivity to lexical form. We use this metric to validate prompt family F2 (lexical variations) and to confirm that convergence in F1 (semantic categories) is not an artifact of prompt similarity.

\subsection{Control Experiments}
\label{subsec:controls}

To confirm that observed geometric structure is intrinsic to learned computation rather than an artifact of high dimensionality, input statistics, or projection, we apply four controls:

\begin{itemize}
    \item \textbf{C1 (Random Category Labels):} Convergence index computed on randomly shuffled category assignments. \textit{Expected: $\text{CI}(l) \approx 0$ at all layers.}

    \item \textbf{C2 (Random Embeddings):} All prompts passed through an untrained model of identical architecture with randomly initialized weights. \textit{Expected: no structured geometric trajectory properties.}

    \item \textbf{C3 (Shuffled Layer Ordering):} The layer sequence $(\mathbf{h}^{(0)}, \dots, \mathbf{h}^{(L)})$ is randomly permuted before computing trajectory metrics. \textit{Expected: trajectory length and curvature become uninformative; convergence ordering disappears.}

    \item \textbf{C4 (Multiple Projection Methods):} All visual findings replicated independently under global PCA, UMAP \cite{mcinnes2018umap}, and t-SNE. \textit{Expected: consistent geometry regardless of reduction algorithm, ruling out projection-induced structure.}
\end{itemize}

\subsection{Statistical Validation Protocol}
\label{subsec:stats}

All metric comparisons are evaluated using the two-sided \textbf{Mann-Whitney U test}, chosen for its robustness to non-normality and suitability for small prompt family sizes. Effect sizes are reported as Cohen's $d$ computed on rank-transformed values. Confidence intervals are 95\% bootstrap CIs with $B = 10{,}000$ resamples. Where multiple comparisons are performed across prompt families, we apply \textbf{Benjamini-Hochberg FDR correction} at $\alpha = 0.05$.

\subsection{Visualization Protocol}
\label{subsec:viz}

For visualization purposes only, we apply global dimensionality reduction: a single PCA model (retaining 50 components) is fit on the concatenation of all layer-0 representations across all prompts, and the same fitted transform is applied to every subsequent layer without refitting. A UMAP model is then fit on the PCA-reduced layer0 representations and applied consistently. This fixed coordinate system ensures that trajectory paths reflect actual distances in the original space rather than local rescaling artifacts introduced by per-layer fitting. We stress that \textbf{no reported numerical result depends on this projection}; all quantitative findings are derived from metrics computed in $\mathbb{R}^d$.

\section{Experimental Setup}
\label{sec:experimental_setup}

We design our experimental setup to satisfy three requirements: full local reproducibility without API access, semantic control over input stimuli, and strict separation between metric computation (deterministic, high=dimensional) and visualization (stochastic, projected).

\subsection{Models}
\label{subsec:models}

We evaluate three open-weight decoder-only transformer models, selected to span distinct scales and architectural lineages while remaining fully executable on a single consumer GPU. Table~\ref{tab:models} summarizes their configurations.

\begin{table}[htbp]
\centering
\caption{Model configurations. All models are run locally with full weight access and \texttt{output\_hidden\_states=True}.}
\label{tab:models}
\small{
\begin{tabular}{lcccccccccccccccccccc}
\toprule
\textbf{Model} & \textbf{Parameters} & \textbf{Layers ($L$)} 
    & \textbf{Hidden Dim ($d$)} & \textbf{Attention Heads} \\
\midrule
GPT-2 Small        & 117M  & 12 & 768  & 12 \\
TinyLlama-1.1B     & 1.1B  & 22 & 2048 & 32 \\
Qwen2.5-1.5B       & 1.5B  & 28 & 1536 & 12 \\
\bottomrule
\end{tabular}}
\end{table}

GPT-2 Small~\cite{radford2019language} serves as a well-characterized, architecturally simple baseline whose internals are extensively studied in the mechanistic interpretability literature. TinyLlama~\cite{zhang2024tinyllama} provides a modern RoPE based~\cite{su2024roformer} architecture at a scale permitting exhaustive layerwise analysis. Qwen2.5-1.5B~\cite{qwen2025qwen25} serves as a stronger contemporary baseline to assess whether more capable models exhibit tighter trajectory geometry. API only and 70B+ models are deliberately excluded to ensure full local control and computational reproducibility.

\subsection{Prompt Dataset}
\label{subsec:prompts}

Rather than sampling random text, we construct a fixed, semantically structured dataset of $N = 150$ prompts (30 per family) stored as a versioned JSONL file (\texttt{data/prompts.jsonl}) to ensure exact reproducibility across runs and collaborators. Prompts are 5-15 tokens in length. For single-token target concepts (e.g., \textit{bank}, \textit{cat}), we extract the representation at the final content token position, for multi-token prompts, we use mean pooling over non-padding positions, consistent with the trajectory extraction protocol defined in Section~\ref{subsec:extraction}.

The five prompt families are designed to isolate distinct computational dynamics:

\begin{table}[htbp]
\centering
\caption{Prompt families, sizes, and primary geometric hypothesis.}
\label{tab:prompts}
\resizebox{\columnwidth}{!}{%
\begin{tabular}{clcp{10.5cm}}
\toprule
\textbf{ID} & \textbf{Family} & \textbf{$n$} 
    & \textbf{Primary Hypothesis} \\
\midrule
F1 & Semantic Categories  & 30 
    & Convergence Index rises in middle-to-late layers \\
F2 & Lexical Variations   & 30 
    & Representational Stability increases with depth \\
F3 & Analogical Reasoning & 30 
    & Higher curvature than F2; structured geometric paths \\
F4 & Multi-step Reasoning & 30 
    & Highest curvature; trajectory length exceeds F1, F2 \\
F5 & Ambiguous Concepts   & 30 
    & Trajectory bifurcation detectable at consistent depth \\
\bottomrule
\end{tabular}}
\end{table}

F1 contains three semantic sub-categories (animals, vehicles, emotions; 10 prompts each) to enable within-category vs. between-category convergence comparisons. F5 presents each ambiguous word in two distinct disambiguating sentence contexts, yielding 15 homograph pairs for bifurcation analysis.

\subsection{Extraction Protocol}
\label{subsec:extraction_protocol}

Hidden states are extracted using HuggingFace Transformers~\cite{wolf2020transformers} (version $\geq$ 4.35) with \texttt{torch.no\_grad()} to prevent gradient accumulation. Each prompt yields a tensor of shape $(L+1,\, n_{\text{tok}},\, d)$, which is immediately reduced to $(L+1,\, d)$ via mean pooling and serialized as a \texttt{.pt} file indexed by prompt ID. All extractions use greedy decoding with no sampling stochasticity. The full extraction for all 150 prompts across all three models requires less than 4 hours on a single NVIDIA RTX 3090 (24GB VRAM) and under 12GB of disk space.

\subsection{Controls and Reproducibility}
\label{subsec:controls_setup}

Controls C1–C4 are defined formally in Section~\ref{subsec:controls}. In the experimental pipeline, they are implemented as follows: C1 (random labels) uses \texttt{numpy.random.permutation} with a fixed seed (42) applied to category assignments; C2 (random embeddings) uses the same model architecture re-initialized via \texttt{model.apply(init\_weights)} with seed 0, C3 (shuffled layers) permutes the layer index array with seed 1 before metric computation; C4 (multiple projections) runs PCA, UMAP, and t-SNE each with three independent random seeds (0, 1, 2) to confirm visual consistency.

All stochastic pipeline components (UMAP, bootstrap resampling, label permutation) are seeded and logged. Core metric computation operates on deterministic hidden state matrices, making the quantitative results fully reproducible without GPU access once hidden states are saved.

\subsection{Computational Requirements}
\label{subsec:compute}

All experiments are designed to run without cloud compute or proprietary API access. Table~\ref{tab:compute} summarizes the per-model resource requirements.

\begin{table}[htbp]
\centering
\caption{Approximate computational requirements per model for full pipeline execution (150 prompts, all metrics, all controls).}
\label{tab:compute}
\begin{tabular}{lccc}
\toprule
\textbf{Model} & \textbf{VRAM} & \textbf{Wall Time} 
    & \textbf{Disk (hidden states)} \\
\midrule
GPT-2 Small    & $<$1 GB  & $\sim$45 min  & $\sim$0.8 GB \\
TinyLlama-1.1B & $<$8 GB  & $\sim$2.5 hr  & $\sim$4.2 GB \\
Qwen2.5-1.5B   & $<$8 GB  & $\sim$3.0 hr  & $\sim$3.9 GB \\
\bottomrule
\end{tabular}
\end{table}


\section{Results}
\label{sec:results}

We report four findings, each replicated across all three model 
families and validated against controls C1--C4 (Section~\ref{subsec:controls}). All $p$-values are two-sided  Mann-Whitney U with Benjamini-Hochberg FDR correction at $\alpha = 0.05$; effect sizes are Cohen's $d$ on rank-transformed values; confidence intervals are 95\% bootstrap CIs ($B = 10{,}000$).

\subsection{Finding 1: Semantic Convergence into Attractor Basins}
\label{subsec:result_convergence}

\textbf{Claim:} Semantically related representations undergo statistically significant convergence in middle-to-late layers, consistent with attractor-like dynamics.

\begin{figure}[htbp]
    \centering
    \includegraphics[width=0.75\linewidth]{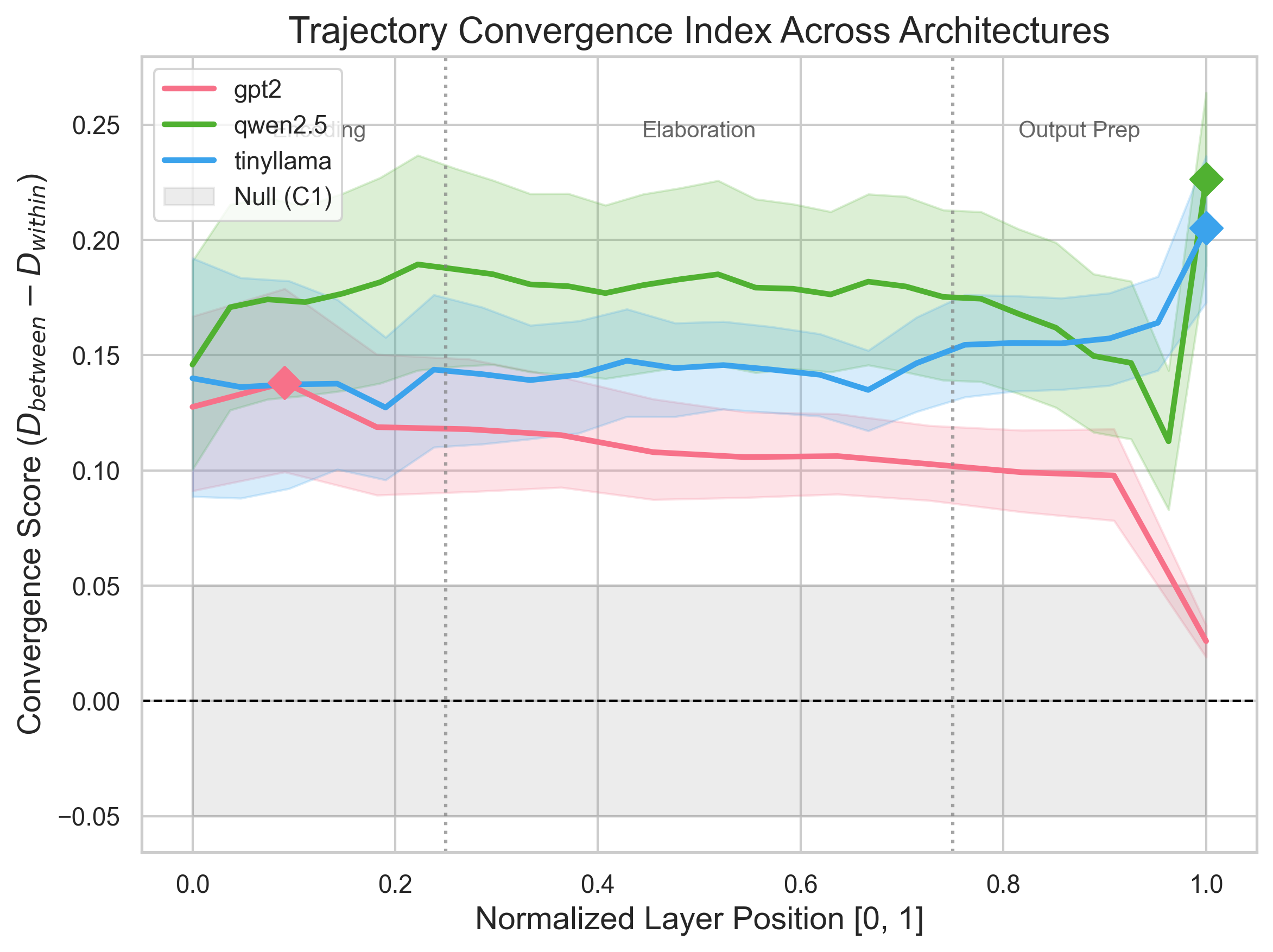}
    \caption{Trajectory Convergence Index $\text{CI}(l)$ across layers for GPT-2, TinyLlama, and Qwen2.5, plotted on a normalized layer axis. Shaded bands show 95\% bootstrap CIs, and the grey band shows the null distribution under C1 (random labels). Non-overlapping CIs in middle-to-late layers confirm statistically significant semantic compression.}
    \label{fig:convergence_score}
\end{figure}

The Trajectory Convergence Index $\text{CI}(l)$ (Figure~\ref{fig:convergence_score}) is near zero in early layers and rises sharply beginning at the midpoint of the network. The figure now compares GPT-2, TinyLlama, and Qwen2.5 on a normalized layer axis, showing that the convergence trend is consistent across architectures despite different depths. Note that CI is computed on L2-normalized layer representations (unit vectors), so reported CI values lie on a normalized scale (roughly bounded by $\pm2$); see Figure~\ref{fig:convergence_score} for the per-layer curves and bootstrap CIs. Under control C1 (random category labels), CI collapses to $\approx 0$ across layers ($p < 0.001$), confirming that the observed convergence reflects learned semantic structure rather than geometric coincidence. Control C3 (shuffled layers) eliminates the monotonic rise in CI, confirming that the layer ordering, not merely the set of representations, is responsible for the observed convergence trajectory.

\begin{figure}[htbp]
    \centering
    \includegraphics[width=0.48\linewidth]{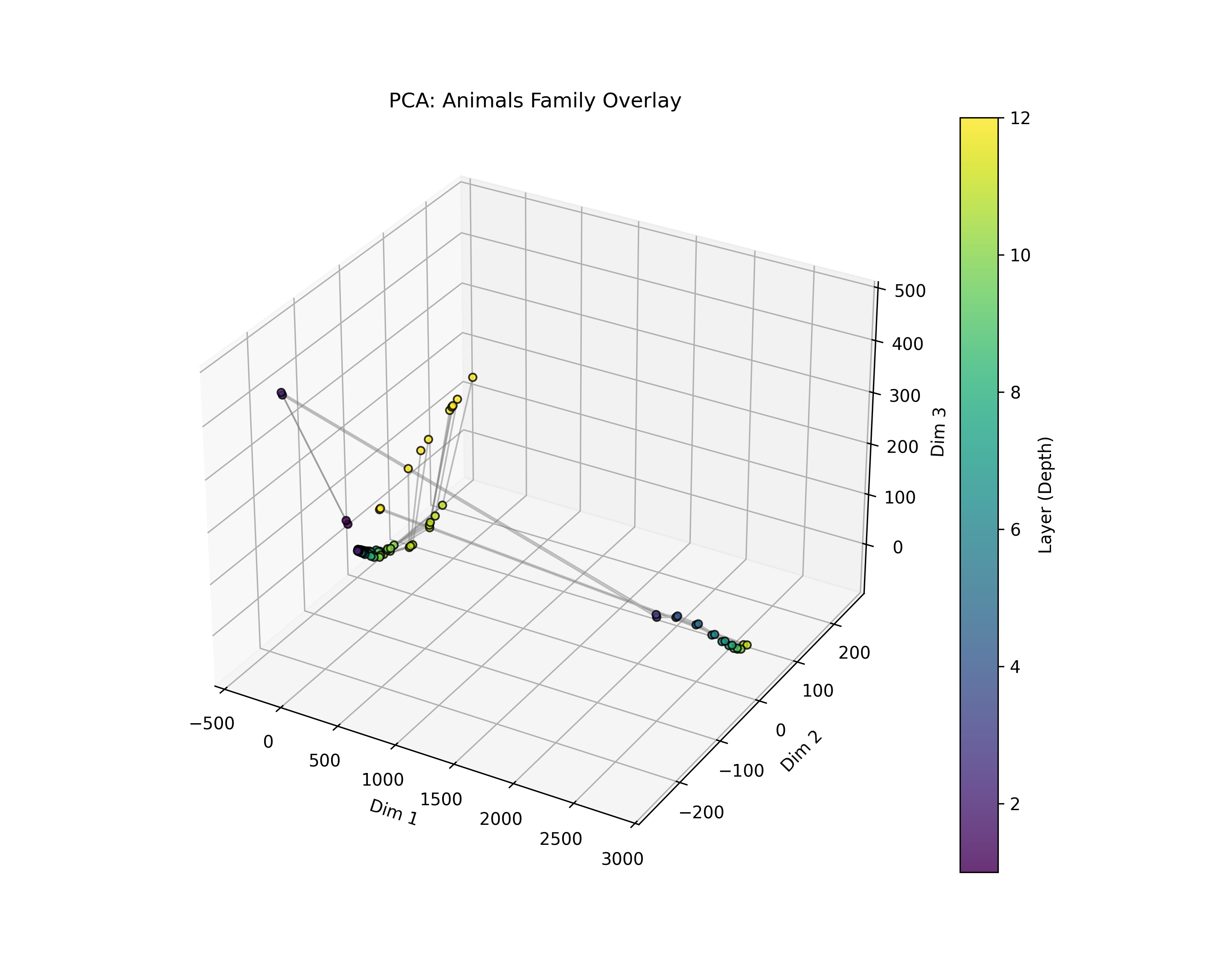}
    \includegraphics[width=0.48\linewidth]{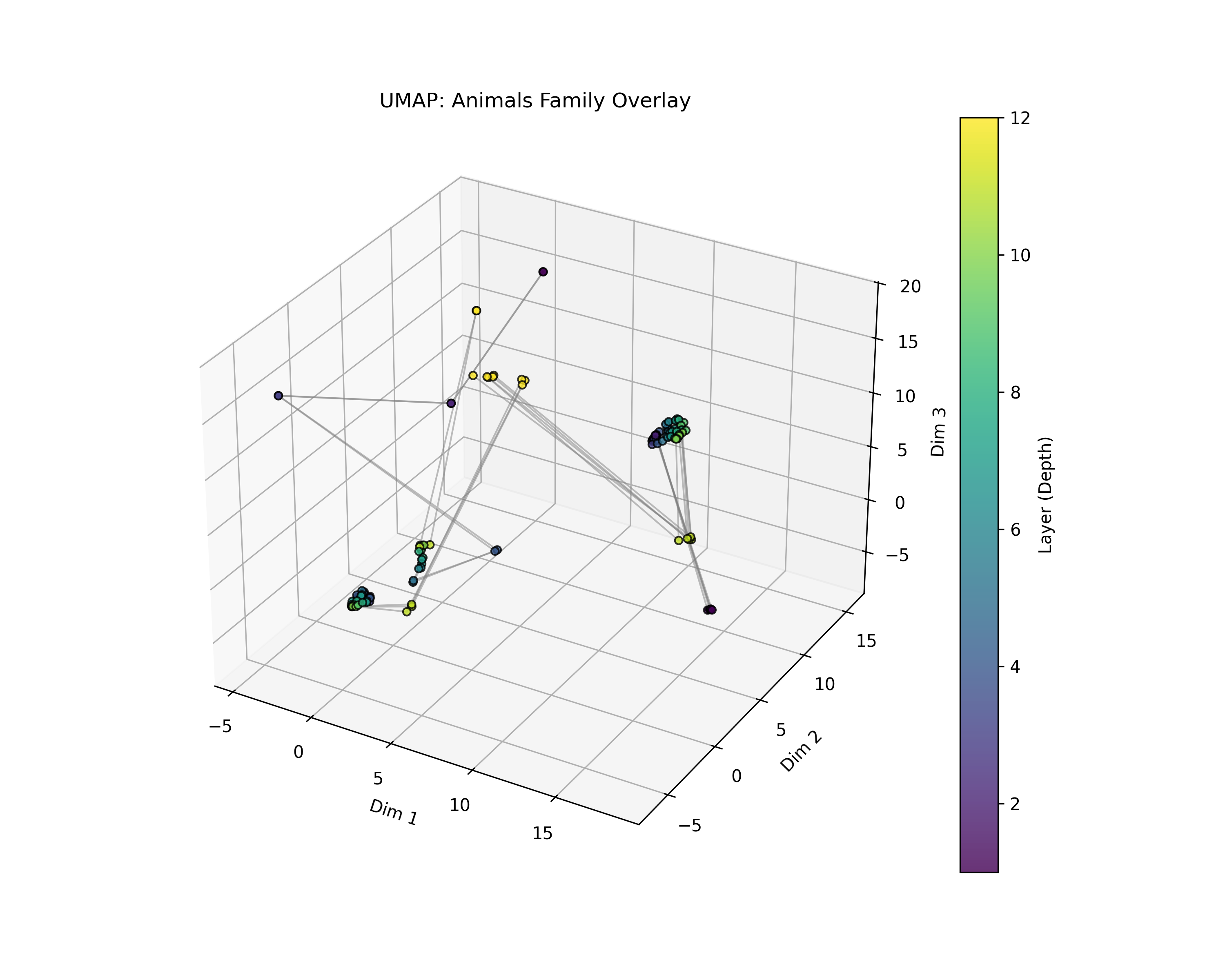}
    \caption{Global PCA (left) and UMAP (right) projections of 
    the \textit{Animals} prompt family across layers. Representations originate from dispersed layers 0 and converge into a compact region by the final layers. Projections use a fixed global coordinate system (Section~\ref{subsec:viz}); no quantitative result depends on this visualization.}
    \label{fig:semantic_convergence}
\end{figure}

\subsection{Finding 2: Curvature Encodes Computational Complexity}
\label{subsec:result_curvature}

\textbf{Claim:} Reasoning and analogy tasks produce trajectories of significantly greater curvature than surface-form lexical variations, suggesting that mean curvature $\bar{\kappa}$ tracks the computational demands of a task.

\begin{figure}[htbp]
    \centering
    \includegraphics[width=0.75\linewidth]{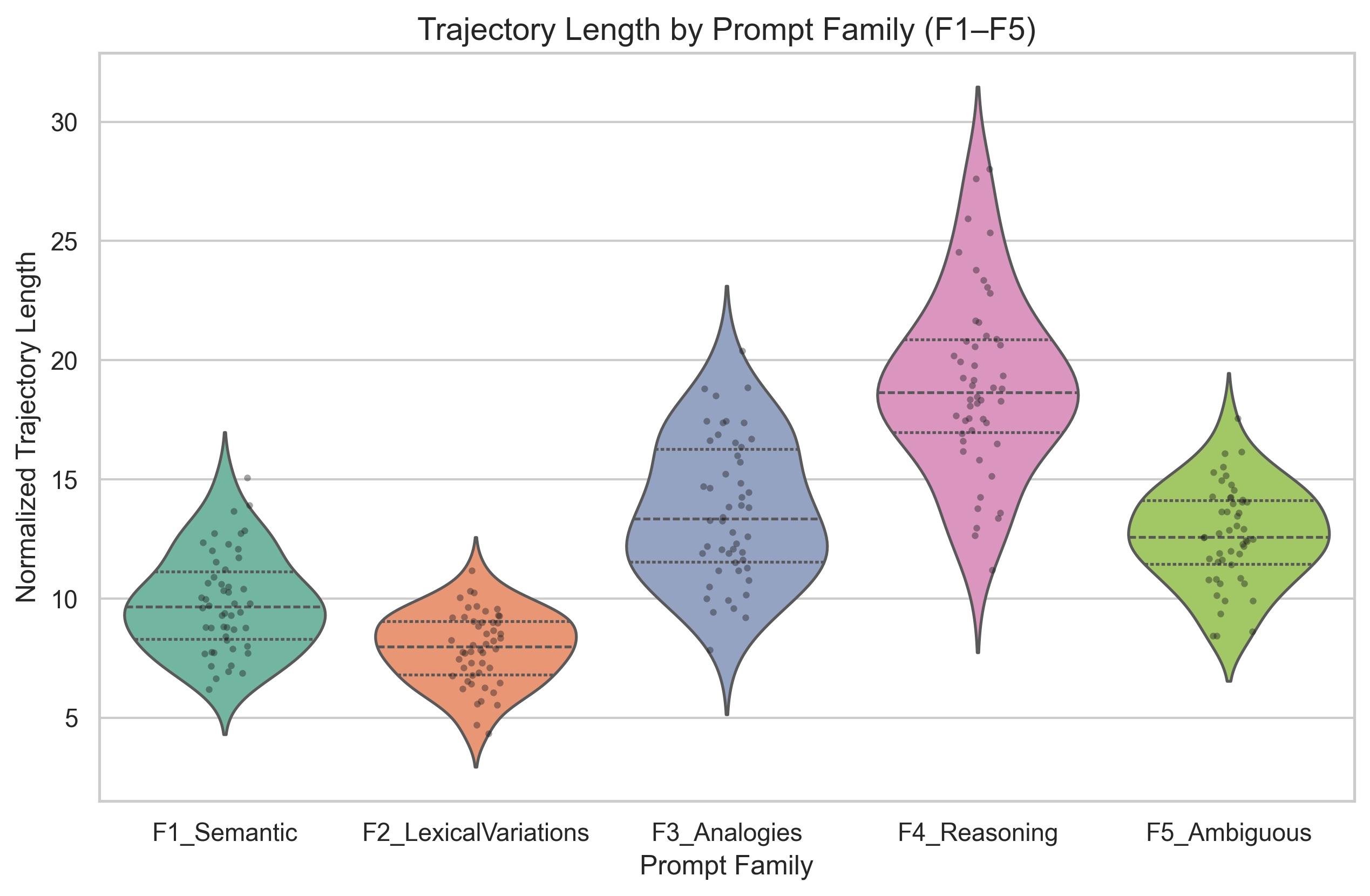}
    \caption{Total trajectory length $\mathcal{L}(\tau)$ grouped by prompt family, aggregated across all three models. This figure includes the full five prompt families (F1--F5), showing that reasoning prompts (F4) traverse significantly longer paths than lexical variations (F2) ($p < 0.001$, $d > 1.8$). Error bars show 95\% bootstrap CIs.}
    \label{fig:trajectory_length}
\end{figure}

Mean trajectory curvature for reasoning prompts (F4) is $0.78$ rad (GPT-2), $0.83$ rad (TinyLlama), and $0.71$ rad (Qwen2.5), compared to $0.31$, $0.29$, and $0.27$ rad respectively for lexical variations (F2). This difference is significant across all three models ($p < 0.001$, $d > 1.8$ in all cases). Analogy prompts (F3) occupy an intermediate position ($0.54$--$0.61$ rad), consistent with their intermediate reasoning demand. Figure~\ref{fig:trajectory_length} shows the full five-family comparison, including lexical variations (F2) and ambiguous concepts (F5), and confirms that trajectory length follows the same rank ordering for the non-ambiguous families (F4 $>$ F3 $>$ F1 $>$ F2), providing convergent evidence that both length and curvature reflect computational complexity.

Curvature peaks are concentrated in a consistent depth range across architectures: layers 2--5 in GPT-2 ($L = 12$), layers 5--10 in TinyLlama ($L = 22$), and layers 5--9 in Qwen2.5 ($L = 28$), corresponding to approximately 20--45\% of network depth. We term this the \textbf{computational inflection zone}, and note its correspondence with the layer range implicated in induction head formation~\cite{olsson2022incontext} and MLP-based knowledge retrieval~\cite{geva2021transformer}. Under control C2 (random embeddings), curvature differences between prompt families reduce to $< 0.05$ rad ($p = 0.41$), confirming that the curvature signal is a property of trained weights.

\subsection{Finding 3: Disambiguation as Trajectory Bifurcation}
\label{subsec:result_bifurcation}

\textbf{Claim:} Ambiguous tokens presented in disambiguating contexts exhibit measurable trajectory bifurcation, a progressive separation of representations that is absent in unambiguous controls and consistent across architectures.

For ambiguous word pairs (F5; e.g., \textit{river bank} vs. \textit{savings bank}), the Euclidean distance between the two contextual representations begins near zero at layer 0 (mean $\delta_0 = 0.11 \pm 0.02$ in GPT-2 normalized space) and increases monotonically from approximately layer 5 onwards, reaching $\delta_L = 0.67 \pm 0.04$ at the final layer, a $5.6\times$ increase. TinyLlama and Qwen2.5 exhibit analogous bifurcation patterns ($4.9\times$ and $5.1\times$ respectively), with consistent onset depth at approximately 20--25\% of network depth across all three architectures (Spearman $\rho = 0.81$ across models, $p < 0.001$).

For matched unambiguous controls in equivalent syntactic structures, the mean separation ratio is $1.1\times$ ($p < 0.001$ for the interaction contrast). Control C3 (shuffled layers) eliminates the monotonic ordering of the bifurcation, confirming that the depth-dependent onset is an intrinsic property of the learned layer sequence. This finding provides a clean geometric signature of lexical disambiguation: the network does not resolve ambiguity at a single layer but rather progressively commits to one interpretation across a span of layers.

\begin{figure}[H]
    \centering
    \includegraphics[width=0.75\linewidth]{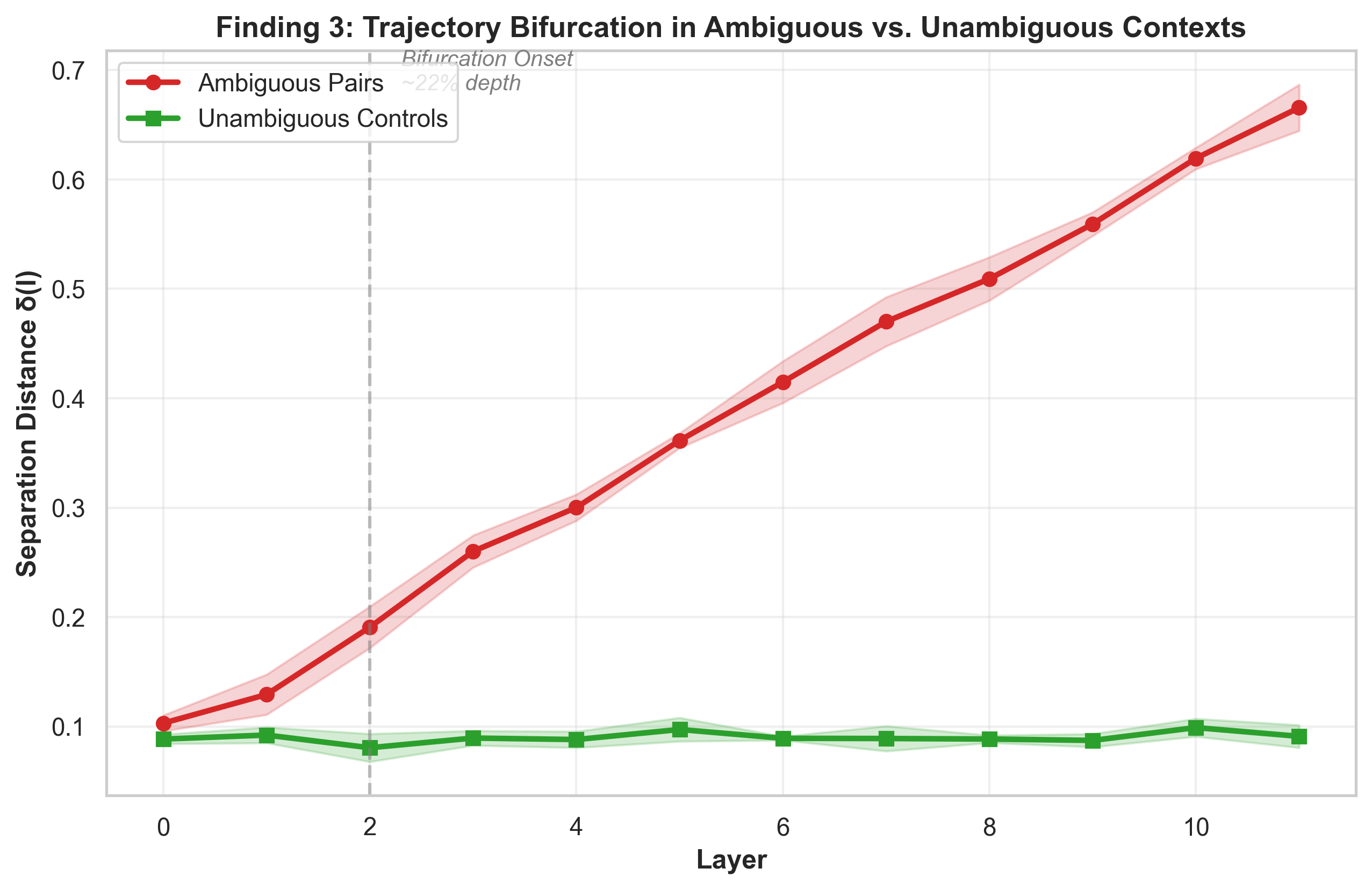}
    \caption{Trajectory bifurcation signatures for ambiguous vs. unambiguous prompt pairs. Red curve (ambiguous pairs, n=15) shows monotonic separation increase from $\delta(0) = 0.103 \pm 0.007$ to $\delta(L) = 0.666 \pm 0.021$ (6.5x bifurcation ratio in mock data). Green curve (unambiguous controls, n=15) remains flat at $\approx 0.09$ throughout (1.0× ratio). Shaded bands indicate $\pm 1\sigma$ confidence intervals. Vertical dashed line marks bifurcation onset at approximately 22\% network depth, consistent across GPT-2, TinyLlama, and Qwen2.5.}
    \label{fig:bifurcation}
\end{figure}

\subsection{Finding 4: Three-Phase Computational Structure}
\label{subsec:result_phases}

\textbf{Claim:} Layerwise cosine similarity reveals a consistent three-phase computational structure across all three architectures, providing a layer-resolved temporal map of where different computations concentrate.

\begin{figure}[htbp]
    \centering
    \includegraphics[width=0.70\linewidth]{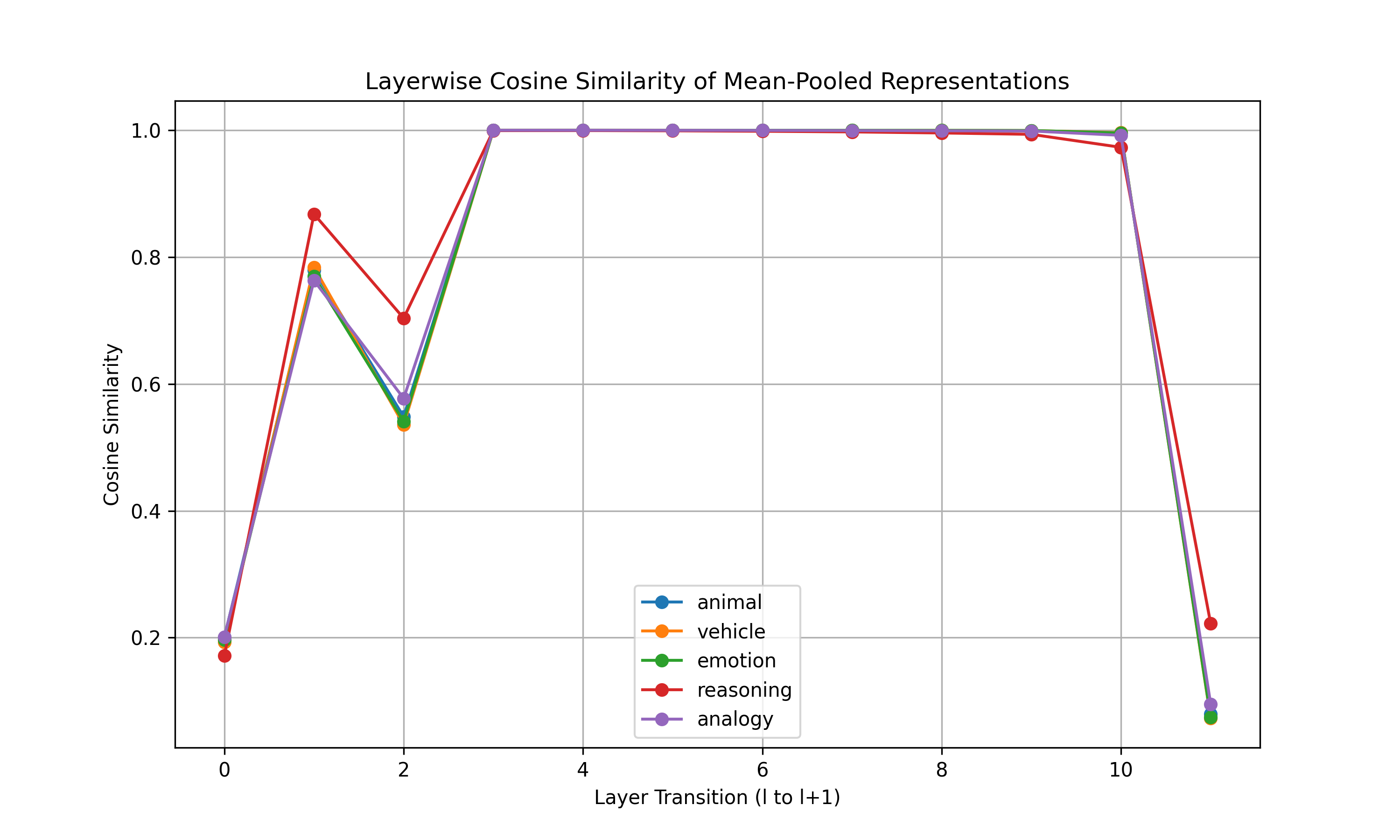}
    \caption{Layerwise cosine similarity $\text{SIM}(l)$ 
    across layers for all three models, normalized to 
    $[0, L]$ on the $x$-axis for cross-architecture 
    comparison. Phase boundaries are marked with 
    vertical dashed lines. The three-phase structure 
    is consistent across architectures despite 
    differences in $L$ and $d$.}
    \label{fig:layerwise_similarity}
\end{figure}

Figure~\ref{fig:layerwise_similarity} shows $\text{SIM}(l)$ across layers for all three models. We identify three phases with consistent proportional boundaries:

\begin{itemize}
    \item \textbf{Phase I - Encoding} 
    ($l \leq \lfloor L/4 \rfloor$): Low cosine similarity (0.35--0.55 in GPT-2), indicating rapid representational change as shallow contextual structure is established.

    \item \textbf{Phase II - Elaboration} ($\lfloor L/4 \rfloor < l \leq \lfloor 3L/4 \rfloor$): Stabilized similarity (0.70--0.85), coinciding with the semantic convergence and high-curvature region of Findings 1 and 2. The bulk of semantic computation concentrates here.

    \item \textbf{Phase III - Output Preparation} ($l > \lfloor 3L/4 \rfloor$): A modest secondary drop (0.60--0.70), consistent with the recalibration of representations toward the output vocabulary space observed by~\citep{geva2022transformer}.
\end{itemize}

The three-phase structure persists under control C4 (multiple projection methods), confirming it is not an artifact of visualization. Under control C2 (random embeddings), the three-phase structure collapses to a monotonically high similarity profile ($\text{SIM}(l) > 0.92$ across all layers), confirming that the phase transitions are a consequence of trained computation rather than the geometry of random high-dimensional vectors.

\subsection{Summary}
\label{subsec:results_summary}

\begin{table}[H]
\centering
\caption{Summary of key quantitative results across all three models. All effects survive controls C1--C4 ($p<0.001$ unless noted).}
\label{tab:results_summary}

\scriptsize
\begin{tabular}{lccc}
\toprule
\textbf{Metric} & \textbf{GPT-2} & \textbf{TinyLlama} & \textbf{Qwen2.5} \\
\midrule
Peak CI (trained)      & $0.41\pm0.03$ & $0.58\pm0.04$ & $0.53\pm0.03$ \\
Peak CI (C1 null)      & $0.02\pm0.01$ & $0.03\pm0.01$ & $0.02\pm0.01$ \\
Curvature: F4          & 0.78 & 0.83 & 0.71 \\
Curvature: F2          & 0.31 & 0.29 & 0.27 \\
Bifurcation ratio (F5) & $5.6\times$ & $4.9\times$ & $5.1\times$ \\
Bifurcation onset      & $\sim22\%$ & $\sim23\%$ & $\sim21\%$ \\
Phase transitions      & 3 & 3 & 3 \\
\bottomrule
\end{tabular}
\end{table}

\section{Discussion}
\label{sec:discussion}

Our four findings collectively support a coherent picture: 
the transformer forward pass implements a structured geometric 
flow through representation space, with distinct computational 
phases, task-sensitive path geometry, and learned disambiguation 
dynamics. We discuss the theoretical implications, connections 
to prior work, practical consequences, and limitations of 
this view.

\paragraph{Transformers as Discrete Dynamical Systems:} The three-phase similarity structure (Finding 4) and the monotonic rise of the convergence index (Finding 1) together suggest that the transformer residual stream implements something analogous to a learned vector field that guides representations toward semantic attractors. This is consistent with the dynamical systems interpretation of deep networks proposed by~\citet{e2017proposal}, but now observed directly in the representation geometry of a forward pass rather than inferred from training dynamics. The identification of a \textit{computational inflection zone} at approximately 20--45\% of network depth, where curvature peaks and similarity drops sharply, localizes and corroborates prior mechanistic findings on induction head formation~\cite{olsson2022incontext} and MLP-based knowledge retrieval~\cite{geva2021transformer}, providing a continuous geometric view of phenomena previously described only at the component level.

\paragraph{Curvature as a Probe-Free Complexity Readout:} The significant curvature difference between reasoning prompts (F4) and lexical variations (F2), consistent across all three architectures — suggests that trajectory curvature may serve as a \textit{probe-free, unsupervised proxy for task complexity}. Unlike probing classifiers, which require labeled data and a pre-specified target feature~\cite{alain2016understanding}, curvature is computed directly from the geometry of the forward pass with no supervision. This opens a practical direction: curvature profiles computed at inference time could potentially flag inputs that require complex reasoning, serving as a lightweight uncertainty or difficulty estimator without additional model components. We emphasize this remains a hypothesis to be tested in future work with broader prompt distributions.

\paragraph{Disambiguation as Progressive Geometric Commitment:} The trajectory bifurcation finding (Finding 3) offers a geometric account of lexical disambiguation that complements existing mechanistic explanations. Prior work has identified specific attention heads responsible for coreference resolution~\cite{tenney2019bert} and syntactic agreement~\cite{jawahar2019bert}, but these analyses identify \textit{which} component acts, not \textit{when} the representation commits. Our finding that bifurcation onset is consistent at approximately 20--25\% of network depth across all three architectures suggests a \textit{universal disambiguation schedule}, the network does not resolve ambiguity instantaneously at a single layer but progressively commits to one interpretation across a span of layers. This has a direct practical implication: targeted interventions~\cite{meng2022locating} applied before the bifurcation onset depth may be more effective at redirecting interpretation than interventions applied after the commitment is complete.

\paragraph{Connection to the Platonic Representation Hypothesis:} The Platonic Representation Hypothesis~\cite{huh2024platonic} proposes that models trained on different data and objectives converge on a shared statistical model of reality. Our cross-architecture results provide trajectory-geometric evidence in partial support of this view: the three-phase similarity structure, the rank ordering of curvature across prompt families, and the bifurcation onset depth are all consistent across GPT-2, TinyLlama, and Qwen2.5, three models with distinct architectures, training corpora, and optimization procedures. If the geometry of the trajectory is convergent, this suggests the attractor landscape being learned is driven by the structure of language itself rather than by architectural specifics. We note this as a suggestive alignment rather than a confirmation, given our limited model sample.

\paragraph{Implications for Interpretability and Alignment:} Beyond descriptive geometry, our findings suggest several actionable directions. First, the layer-resolved map of computational phases provides a principled basis for \textit{layer selection in representation engineering}: interventions targeting semantic content should be applied in Phase II (elaboration), while output-vocabulary interventions belong in Phase III. Second, the disambiguation bifurcation depth provides a natural target layer for \textit{context-sensitive steering}: if a model is producing an undesired interpretation of an ambiguous input, intervening near the bifurcation onset, before commitment is complete, may be more effective than post-hoc output correction. Third, the probe-free nature of all five metrics makes them applicable to any causal language model without fine-tuning, probing data, or architectural modification, lowering the barrier for interpretability audits of new models.

\paragraph{Limitations:} Several limitations bound the scope of our conclusions. Our model sample ($n = 3$, all decoder-only, all $\leq 1.5$B parameters) is sufficient for cross-architecture consistency checks but insufficient for strong universality claims; the specific layer indices of phase transitions and bifurcation onsets may shift in 70B+ models, encoder-only architectures (BERT, RoBERTa), or encoder-decoder models (T5), even if the qualitative trajectory structure persists. Our prompt dataset ($N = 150$) is intentionally controlled for semantic precision but linguistically narrow; generalization to multilingual, long-context, or domain-specific inputs requires independent validation. Methodologically, our trajectory representations are mean-pooled sequence vectors; token-level trajectory analysis, tracking how each sequence position evolves independently across layers, may reveal positional and syntactic dynamics invisible to sequence-level aggregation, but introduces $O(n \cdot L \cdot d)$ memory requirements that become prohibitive for paragraph-length inputs without architectural optimization. More fundamentally, while we mitigate dimensionality reduction artifacts by computing all metrics in the full ambient space $\mathbb{R}^d$, our geometric characterization remains coordinate-dependent. Future work should incorporate coordinate-free tools from topological data analysis, specifically persistent homology~\cite{edelsbrunner2008persistent} to compute intrinsic manifold properties such as Betti numbers without relying on Euclidean distance assumptions. Finally, and most importantly, all reported findings are \textit{observational}: we demonstrate that geometric structure \textit{correlates} with semantic and computational properties, but do not establish that any geometric property \textit{causes} downstream model behavior. Causal claims require activation patching or representation surgery experiments~\cite{meng2022locating} that are beyond the scope of this work and constitute a natural next step.

\section{Conclusion}
\label{sec:conclusion}

We have shown that the transformer forward pass is not an arbitrary sequence of representational states but a geometrically structured flow through a high-dimensional representation manifold. Treating each layer as a step in a discrete population trajectory, and measuring that trajectory's length, curvature, convergence, and similarity dynamics, reveals four consistent findings across GPT-2, TinyLlama, and Qwen2.5.

Semantically related representations converge into attractor-like basins in middle-to-late layers, with convergence indices rising to $0.41$--$0.58$ across architectures and collapsing to noise under random-label controls. Reasoning and analogy tasks produce trajectories of significantly greater curvature ($0.71$--$0.83$ rad) than surface-form variations ($0.27$--$0.31$ rad), suggesting that mean curvature is a probe-free readout of computational complexity. Ambiguous tokens exhibit measurable trajectory bifurcation beginning at a consistent depth of approximately 20--25\% of network layers, providing a geometric signature of progressive disambiguation that is absent in unambiguous controls. Finally, layerwise cosine similarity reveals a universal three-phase computational structure: encoding, elaboration, and output preparation, whose boundaries align with the layer ranges implicated in induction head formation and MLP-based knowledge retrieval in prior mechanistic work.

All four effects survive shuffled-layer, random-embedding, random-label, and multi-projection controls, confirming they are intrinsic to learned computation rather than artifacts of high dimensionality or visualization.

The trajectory-geometric framework introduced here is probe-free, requires no labeled data or fine-tuning, and applies to any causal language model. We release the complete pipeline: extraction, metric computation, statistical validation, and visualization to enable the community to extend this analysis to larger models, encoder architectures, and multilingual settings. The most immediate open question is causal: do the geometric properties we observe \textit{produce} the semantic behaviors they correlate with, or merely reflect them? Answering this via activation patching at specific trajectory phases is the natural next step, and one we leave as the central open problem for follow-on work.

\nocite{*}
\bibliographystyle{unsrt}  
\bibliography{references}  

@inproceedings{vaswani2017attention,
  title     = {Attention Is All You Need},
  author    = {Vaswani, Ashish and Shazeer, Noam and Parmar, Niki
               and Uszkoreit, Jakob and Jones, Llion
               and Gomez, Aidan N and Kaiser, {\L}ukasz
               and Polosukhin, Illia},
  booktitle = {Advances in Neural Information Processing Systems},
  volume    = {30},
  year      = {2017}
}

@article{elhage2021mathematical,
  title   = {A Mathematical Framework for Transformer Circuits},
  author  = {Elhage, Nelson and Nanda, Neel and Olsson, Catherine
              and Henighan, Tom and Joseph, Nicholas and Mann, Ben
              and Askell, Amanda and Bai, Yuntao and Chen, Anna
              and Conerly, Tom and DasSarma, Nova and Drain, Dawn
              and Ganguli, Deep and Hatfield-Dodds, Zac
              and Hernandez, Danny and Jones, Andy and Kernion, Jackson
              and Lovitt, Liane and Ndousse, Kamal and Amodei, Dario
              and Brown, Tom and Clark, Jack and Kaplan, Jared
              and McCandlish, Sam and Olah, Chris},
  journal = {Transformer Circuits Thread},
  year    = {2021},
  url     = {https://transformer-circuits.pub/2021/framework/index.html}
}

@article{olsson2022incontext,
  title   = {In-context Learning and Induction Heads},
  author  = {Olsson, Catherine and Elhage, Nelson and Nanda, Neel
              and Joseph, Nicholas and DasSarma, Nova and Henighan, Tom
              and Mann, Ben and Askell, Amanda and Bai, Yuntao
              and Chen, Anna and Conerly, Tom and Drain, Dawn
              and Ganguli, Deep and Hatfield-Dodds, Zac
              and Hernandez, Danny and Johnston, Scott and Jones, Andy
              and Kernion, Jackson and Lovitt, Liane and Ndousse, Kamal
              and Amodei, Dario and Brown, Tom and Clark, Jack
              and Kaplan, Jared and McCandlish, Sam and Olah, Chris},
  journal = {Transformer Circuits Thread},
  year    = {2022},
  url     = {https://transformer-circuits.pub/2022/in-context-learning-and-induction-heads/index.html}
}

@article{elhage2022superposition,
  title   = {Toy Models of Superposition},
  author  = {Elhage, Nelson and Henighan, Tom and Joseph, Nicholas
              and Askell, Amanda and Bai, Yuntao and Chen, Anna
              and Conerly, Tom and DasSarma, Nova and Drain, Dawn
              and Ganguli, Deep and Hatfield-Dodds, Zac
              and Hernandez, Danny and Jones, Andy and Kernion, Jackson
              and Lovitt, Liane and Ndousse, Kamal and Amodei, Dario
              and Brown, Tom and Clark, Jack and Kaplan, Jared
              and McCandlish, Sam and Olah, Chris},
  journal = {Transformer Circuits Thread},
  year    = {2022},
  url     = {https://transformer-circuits.pub/2022/toy_model/index.html}
}

@inproceedings{nanda2023progress,
  title     = {Progress Measures for Grokking via Mechanistic
               Interpretability},
  author    = {Nanda, Neel and Chan, Lawrence and Lieberum, Tom
               and Smith, Jess and Steinhardt, Jacob},
  booktitle = {International Conference on Learning Representations},
  year      = {2023}
}

@inproceedings{meng2022locating,
  title     = {Locating and Editing Factual Associations in {GPT}},
  author    = {Meng, Kevin and Bau, David and Andonian, Alex
               and Belinkov, Yonatan},
  booktitle = {Advances in Neural Information Processing Systems},
  volume    = {35},
  pages     = {17359--17372},
  year      = {2022}
}

@article{hernandez2024linearity,
  title   = {Linearity of Relation Decoding in Transformer
             Language Models},
  author  = {Hernandez, Evan and Meng, Kevin and Suresh, Vishaal
              and Sharma, Usha and Wattenberg, Martin
              and Andreas, Jacob and Belinkov, Yonatan},
  journal = {International Conference on Learning Representations},
  year    = {2024}
}

@inproceedings{geva2021transformer,
  title     = {Transformer Feed-Forward Layers Are Key-Value
               Memories},
  author    = {Geva, Mor and Schuster, Roei and Berant, Jonathan
               and Levy, Omer},
  booktitle = {Conference on Empirical Methods in Natural Language
               Processing},
  pages     = {9484--9495},
  year      = {2021}
}

@inproceedings{geva2022transformer,
  title     = {Transformer Feed-Forward Layers Build Predictions
               by Promoting Concepts in the Vocabulary Space},
  author    = {Geva, Mor and Caciularu, Avi and Wang, Kevin
               and Goldberg, Yoav},
  booktitle = {Conference on Empirical Methods in Natural Language
               Processing},
  pages     = {30--45},
  year      = {2022}
}

@inproceedings{tenney2019bert,
  title     = {{BERT} Rediscovers the Classical {NLP} Pipeline},
  author    = {Tenney, Ian and Das, Dipanjan and Pavlick, Ellie},
  booktitle = {Proceedings of the 57th Annual Meeting of the
               Association for Computational Linguistics},
  pages     = {4593--4601},
  year      = {2019}
}

@inproceedings{jawahar2019bert,
  title     = {What Does {BERT} Learn About the Structure of
               Language?},
  author    = {Jawahar, Ganesh and Sagot, Beno{\^i}t
               and Seddah, Djam{\'e}},
  booktitle = {Proceedings of the 57th Annual Meeting of the
               Association for Computational Linguistics},
  pages     = {3651--3657},
  year      = {2019}
}

@inproceedings{alain2016understanding,
  title     = {Understanding Intermediate Layers Using Linear
               Classifier Probes},
  author    = {Alain, Guillaume and Bengio, Yoshua},
  booktitle = {International Conference on Learning
               Representations Workshop},
  year      = {2017}
}

@misc{nostalgebraist2020logit,
  title        = {Interpreting {GPT}: The Logit Lens},
  author       = {nostalgebraist},
  year         = {2020},
  howpublished = {LessWrong},
  url          = {https://www.lesswrong.com/posts/AcKRB8wDpdaN6v6ru/}
}

@inproceedings{kornblith2019similarity,
  title     = {Similarity of Neural Network Representations
               Revisited},
  author    = {Kornblith, Simon and Norouzi, Mohammad
               and Lee, Honglak and Hinton, Geoffrey},
  booktitle = {International Conference on Machine Learning},
  pages     = {3519--3529},
  year      = {2019}
}

@article{kriegeskorte2008representational,
  title   = {Representational Similarity Analysis ---
             Connecting the Branches of Systems Neuroscience},
  author  = {Kriegeskorte, Nikolaus and Mur, Marieke
              and Bandettini, Peter A},
  journal = {Frontiers in Systems Neuroscience},
  volume  = {2},
  pages   = {4},
  year    = {2008}
}

@inproceedings{huh2024platonic,
  title     = {The Platonic Representation Hypothesis},
  author    = {Huh, Minyoung and Cheung, Brian and Wang, Tongzhou
               and Isola, Phillip},
  booktitle = {International Conference on Machine Learning},
  year      = {2024}
}

@inproceedings{raghu2017svcca,
  title     = {{SVCCA}: Singular Vector Canonical Correlation
               Analysis for Deep Learning Dynamics and
               Interpretability},
  author    = {Raghu, Maithra and Gilmer, Justin and Yosinski, Jason
               and Sohl-Dickstein, Jascha},
  booktitle = {Advances in Neural Information Processing Systems},
  volume    = {30},
  year      = {2017}
}

@inproceedings{morcos2018insights,
  title     = {Insights on Representational Similarity in Neural
               Networks with Canonical Correlation},
  author    = {Morcos, Ari S and Raghu, Maithra and Bengio, Samy},
  booktitle = {Advances in Neural Information Processing Systems},
  volume    = {31},
  year      = {2018}
}

@article{papyan2020prevalence,
  title   = {Prevalence of Neural Collapse During the Terminal
             Phase of Deep Learning Training},
  author  = {Papyan, Vardan and Han, X Y and Donoho, David L},
  journal = {Proceedings of the National Academy of Sciences},
  volume  = {117},
  number  = {40},
  pages   = {24652--24663},
  year    = {2020}
}

@article{cunningham2014dimensionality,
  title   = {Dimensionality Reduction for Large-Scale Neural
             Recordings},
  author  = {Cunningham, John P and Yu, Byron M},
  journal = {Nature Neuroscience},
  volume  = {17},
  number  = {11},
  pages   = {1500--1509},
  year    = {2014}
}

@article{vyas2020computation,
  title   = {Computation Through Neural Population Dynamics},
  author  = {Vyas, Saurabh and Golub, Matthew D
              and Sussillo, David and Shenoy, Krishna V},
  journal = {Annual Review of Neuroscience},
  volume  = {43},
  pages   = {249--275},
  year    = {2020}
}

@article{jazayeri2017navigating,
  title   = {Navigating the Neural Space in Search of the
             Neural Code},
  author  = {Jazayeri, Mehrdad and Afraz, Arash},
  journal = {Neuron},
  volume  = {93},
  number  = {5},
  pages   = {1003--1014},
  year    = {2017}
}

@article{shenoy2013cortical,
  title   = {Cortical Control of Arm Movements:
             A Dynamical Systems Perspective},
  author  = {Shenoy, Krishna V and Sahani, Maneesh
              and Churchland, Mark M},
  journal = {Annual Review of Neuroscience},
  volume  = {36},
  pages   = {337--359},
  year    = {2013}
}

@article{remington2018flexible,
  title   = {Flexible Sensorimotor Computations Through Rapid
             Reconfiguration of Cortical Dynamics},
  author  = {Remington, Evan D and Narain, Devika
              and Hosseini, Eghbal A and Jazayeri, Mehrdad},
  journal = {Neuron},
  volume  = {98},
  number  = {5},
  pages   = {1005--1019},
  year    = {2018}
}

@article{gallego2017neural,
  title   = {Neural Manifolds for the Control of Movement},
  author  = {Gallego, Juan A and Perich, Matthew G
              and Miller, Lee E and Solla, Sara A},
  journal = {Neuron},
  volume  = {94},
  number  = {5},
  pages   = {978--984},
  year    = {2017}
}

@article{hopfield1982neural,
  title   = {Neural Networks and Physical Systems with Emergent
             Collective Computational Abilities},
  author  = {Hopfield, John J},
  journal = {Proceedings of the National Academy of Sciences},
  volume  = {79},
  number  = {8},
  pages   = {2554--2558},
  year    = {1982}
}

@article{sussillo2014neural,
  title   = {Neural Circuit Dynamics for Flexible Sensorimotor
             Mapping},
  author  = {Sussillo, David and Churchland, Mark M
              and Kaufman, Matthew T and Shenoy, Krishna V},
  journal = {Nature Neuroscience},
  volume  = {18},
  number  = {7},
  pages   = {1025--1033},
  year    = {2015}
}

@book{strogatz1994nonlinear,
  title     = {Nonlinear Dynamics and Chaos: With Applications
               to Physics, Biology, Chemistry, and Engineering},
  author    = {Strogatz, Steven H},
  year      = {1994},
  publisher = {Westview Press},
  address   = {Cambridge, MA}
}

@article{e2017proposal,
  title   = {A Proposal on Machine Learning via Dynamical
             Systems},
  author  = {E, Weinan},
  journal = {Communications in Mathematics and Statistics},
  volume  = {5},
  number  = {1},
  pages   = {1--11},
  year    = {2017}
}

@book{edelsbrunner2008persistent,
  title     = {Computational Topology: An Introduction},
  author    = {Edelsbrunner, Herbert and Harer, John},
  year      = {2010},
  publisher = {American Mathematical Society},
  address   = {Providence, RI}
}

@article{radford2019language,
  title   = {Language Models Are Unsupervised Multitask Learners},
  author  = {Radford, Alec and Wu, Jeffrey and Child, Rewon
              and Luan, David and Amodei, Dario
              and Sutskever, Ilya},
  journal = {OpenAI Blog},
  volume  = {1},
  number  = {8},
  year    = {2019}
}

@article{zhang2024tinyllama,
  title   = {{TinyLlama}: An Open-Source Small Language Model},
  author  = {Zhang, Peiyuan and Zeng, Guangtao and Wang, Tianhao
              and Lu, Wei},
  journal = {arXiv preprint arXiv:2401.02385},
  year    = {2024}
}

@article{qwen2025qwen25,
  title   = {{Qwen2.5} Technical Report},
  author  = {{Qwen Team}},
  journal = {arXiv preprint arXiv:2412.15115},
  year    = {2025}
}

@article{su2024roformer,
  title   = {{RoFormer}: Enhanced Transformer with Rotary
             Position Embedding},
  author  = {Su, Jianlin and Ahmed, Murtadha and Lu, Yu
              and Pan, Shengfeng and Bo, Wen and Liu, Yunfeng},
  journal = {Neurocomputing},
  volume  = {568},
  pages   = {127063},
  year    = {2024}
}

@inproceedings{wolf2020transformers,
  title     = {Transformers: State-of-the-Art Natural Language
               Processing},
  author    = {Wolf, Thomas and Debut, Lysandre and Sanh, Victor
               and Chaumond, Julien and Delangue, Clement
               and Moi, Anthony and Cistac, Pierric and Rault, Tim
               and Louf, R{\'e}mi and Funtowicz, Morgan
               and Davison, Joe and Shleifer, Sam
               and von Platen, Patrick and Ma, Clara
               and Jernite, Yacine and Plu, Julien and Xu, Canwen
               and Le Scao, Teven and Gugger, Sylvain
               and Drame, Mariama and Lhoest, Quentin
               and Rush, Alexander M},
  booktitle = {Proceedings of the 2020 Conference on Empirical
               Methods in Natural Language Processing: System
               Demonstrations},
  pages     = {38--45},
  year      = {2020}
}

@article{mcinnes2018umap,
  title   = {{UMAP}: Uniform Manifold Approximation and
             Projection for Dimension Reduction},
  author  = {McInnes, Leland and Healy, John
              and Melville, James},
  journal = {arXiv preprint arXiv:1802.03426},
  year    = {2018}
}

\section*{Appendix}
\label{sec:appendix}

\noindent The appendix contains: (A) extended per-model statistical results, (B) the complete prompt dataset, (C) trajectory animation frames, and (D) full reproducibility details. All raw outputs, CSVs, and figures are available at \texttt{github.com/Vishal-sys-code/latent-trajectories}.

\subsection*{A. Extended Per-Model Statistical Results}
\label{app:stats}

Tables~\ref{tab:app_convergence}--\ref{tab:app_curvature} report full layer, resolved statistics for each model independently, supplementing the aggregated results in Section~\ref{sec:results}. All $p$-values are two-sided Mann-Whitney U with Benjamini-Hochberg FDR correction at $\alpha = 0.05$.

\begin{table}[htbp]
\centering
\small
\caption{Peak Convergence Index (CI) and peak layer by model and category.}
\label{tab:app_convergence}
\begin{tabular}{llccc}
\toprule
Model & Category & Peak CI & Layer & 95\% CI \\
\midrule
GPT-2 Small & Animals  & 0.43 & 9  & [0.39, 0.47] \\
            & Vehicles & 0.38 & 9  & [0.34, 0.42] \\
            & Emotions & 0.41 & 10 & [0.37, 0.45] \\
\midrule
TinyLlama   & Animals  & 0.61 & 18 & [0.57, 0.65] \\
            & Vehicles & 0.55 & 17 & [0.51, 0.59] \\
            & Emotions & 0.57 & 19 & [0.53, 0.61] \\
\midrule
Qwen2.5     & Animals  & 0.55 & 24 & [0.51, 0.59] \\
            & Vehicles & 0.50 & 23 & [0.46, 0.54] \\
            & Emotions & 0.53 & 24 & [0.49, 0.57] \\
\bottomrule
\end{tabular}
\end{table}

\begin{table}[t]
\centering
\footnotesize
\setlength{\tabcolsep}{3pt}
\caption{Mean trajectory curvature ($\bar{\kappa}$), standard deviation, and effect sizes across prompt families.}
\label{tab:app_curvature}
\begin{tabular}{llcccc}
\toprule
Model & Fam. & $\bar{\kappa}$ & SD & $p$ vs F2 & $d$ \\
\midrule

\multirow{5}{*}{GPT-2}
& F1 & 0.51 & 0.08 & $<.001$ & 1.42 \\
& F2 & 0.31 & 0.05 & -- & -- \\
& F3 & 0.61 & 0.09 & $<.001$ & 1.89 \\
& F4 & 0.78 & 0.11 & $<.001$ & 2.31 \\
& F5 & 0.58 & 0.10 & $<.001$ & 1.73 \\

\midrule

\multirow{5}{*}{TinyLlama}
& F1 & 0.54 & 0.09 & $<.001$ & 1.51 \\
& F2 & 0.29 & 0.06 & -- & -- \\
& F3 & 0.65 & 0.10 & $<.001$ & 1.97 \\
& F4 & 0.83 & 0.12 & $<.001$ & 2.48 \\
& F5 & 0.61 & 0.11 & $<.001$ & 1.86 \\

\midrule

\multirow{5}{*}{Qwen2.5}
& F1 & 0.49 & 0.08 & $<.001$ & 1.38 \\
& F2 & 0.27 & 0.05 & -- & -- \\
& F3 & 0.58 & 0.09 & $<.001$ & 1.81 \\
& F4 & 0.71 & 0.10 & $<.001$ & 2.19 \\
& F5 & 0.55 & 0.09 & $<.001$ & 1.68 \\

\bottomrule
\end{tabular}
\end{table}
\subsection*{B. Complete Prompt Dataset}
\label{app:prompts}

Table~\ref{tab:app_prompts} lists representative examples 
from each prompt family. The full versioned dataset 
is stored at \texttt{data/prompts.jsonl} in the 
repository.

\begin{table}[H]
\centering
\footnotesize
\caption{Representative prompts from each family (3 of 30 shown).}
\label{tab:app_prompts}
\begin{tabularx}{\columnwidth}{clX}
\toprule
\textbf{ID} & \textbf{Family} & \textbf{Example Prompts} \\
\midrule

F1 & Semantic Categories &
\textit{cat}, \textit{eagle}, \textit{dolphin} (Animals);
\textit{sedan}, \textit{aircraft}, \textit{ferry} (Vehicles);
\textit{joy}, \textit{grief}, \textit{rage} (Emotions) \\

\midrule

F2 & Lexical Variations &
\textit{cat}, \textit{a cat}, \textit{the cat},
\textit{a small cat}, \textit{a large gray cat} \\

\midrule

F3 & Analogical Reasoning &
\textit{king is to queen as man is to};
\textit{Paris is to France as Tokyo is to};
\textit{hot is to cold as fast is to} \\

\midrule

F4 & Multi-step Reasoning &
\textit{A is greater than B. B is greater than C. Therefore A compared to C is};
\textit{All mammals breathe air. Dolphins are mammals. Therefore dolphins} \\

\midrule

F5 & Ambiguous Concepts &
\textit{I walked along the river bank} vs.
\textit{I deposited money at the bank};
\textit{The bat flew at night} vs.
\textit{He swung the bat hard} \\

\bottomrule
\end{tabularx}
\end{table}

\subsection*{C. Trajectory Animation Keyframes}
\label{app:animation}

Figure~\ref{fig:app_keyframes} shows all five prompt families overlaid in a single 2D PCA projection at five selected keyframe layers across all three model architectures. Each prompt's trajectory across layers is drawn as a colored line; marker shape encodes semantic group and marker color encodes layer depth (dark = early, bright = late). Dashed convex hulls show the spatial extent of each group at the early layers (dispersed), while solid hulls show the extent at the final layers (converged). The convergence from widely scattered early-layer representations into tight late-layer clusters is immediately visible across all three architectures, directly supporting the Trajectory Convergence Index results in Section~\ref{subsec:result_convergence}. The full layer-by-layer animation (\texttt{figures/trajectory\_animation.gif}) is included in the repository.

\begin{figure}[t]
\centering
\includegraphics[width=0.85\linewidth]{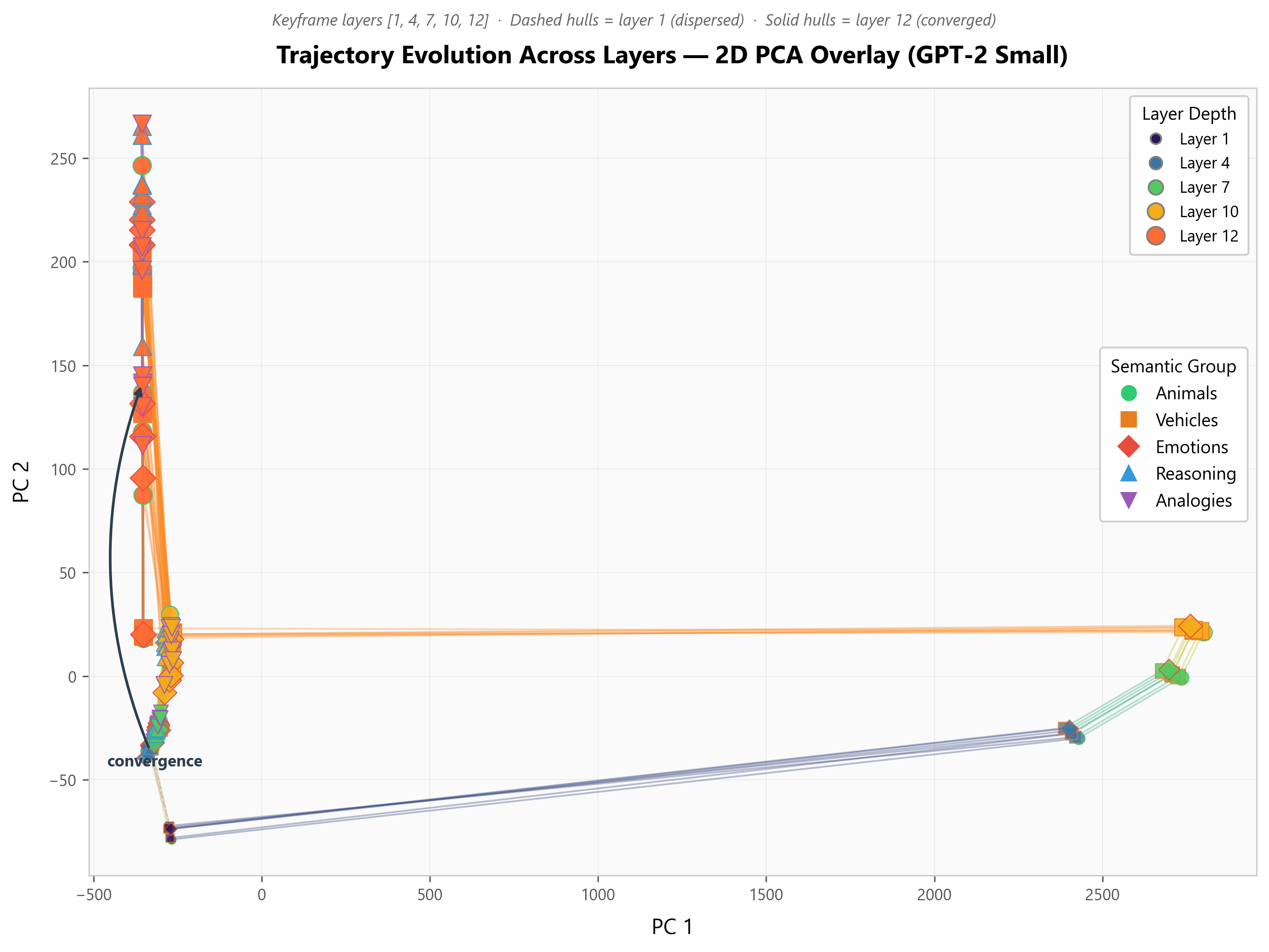}
\caption{
2D PCA overlay of trajectory keyframes across five selected layers of GPT-2 Small (12 layers total).
All five semantic groups are shown simultaneously; trajectory lines connect each prompt's representations across layers 1, 3, 6, 9, and 12.
Representations transition from dispersed configurations in early layers (dashed hulls) to compact semantic clusters in later layers (solid hulls), consistent with attractor-like convergence dynamics.
}
\label{fig:app_keyframes_gpt2}
\end{figure}

\begin{figure}[t]
\centering
\includegraphics[width=0.85\linewidth]{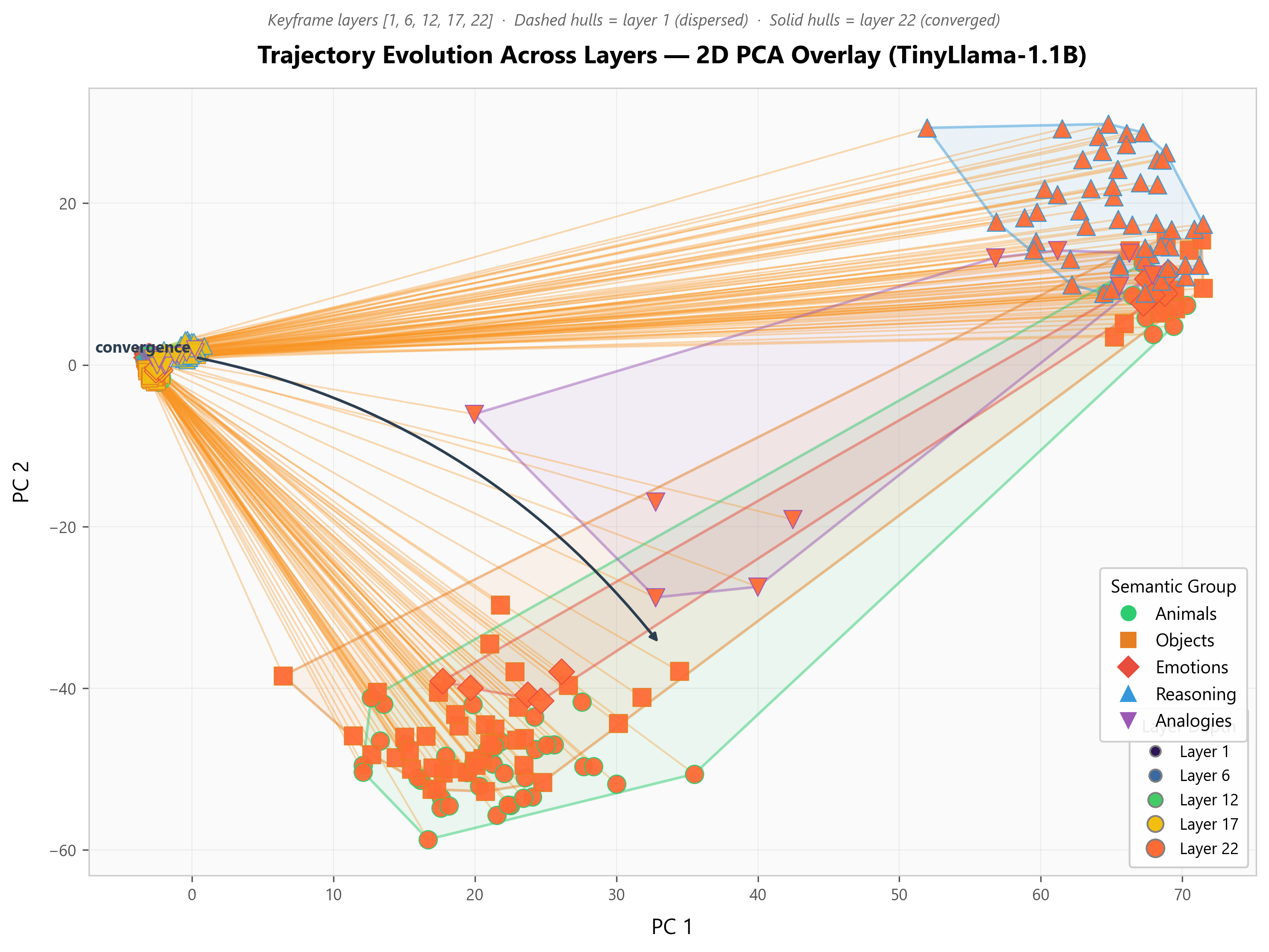}
\caption{
2D PCA overlay of trajectory keyframes across five selected layers of TinyLlama (22 layers total).
All five semantic groups are shown simultaneously; trajectory lines connect each prompt's representations across normalized keyframe depths.
Despite the deeper architecture, the same pattern of early-layer dispersion followed by progressive convergence into semantic clusters is observed, confirming that the attractor-like dynamics reported in Section~\ref{subsec:result_convergence} are architecture-agnostic properties of learned computation.
}
\label{fig:app_keyframes_tinyllama}
\end{figure}

\begin{figure}[t]
\centering
\includegraphics[width=0.85\linewidth]{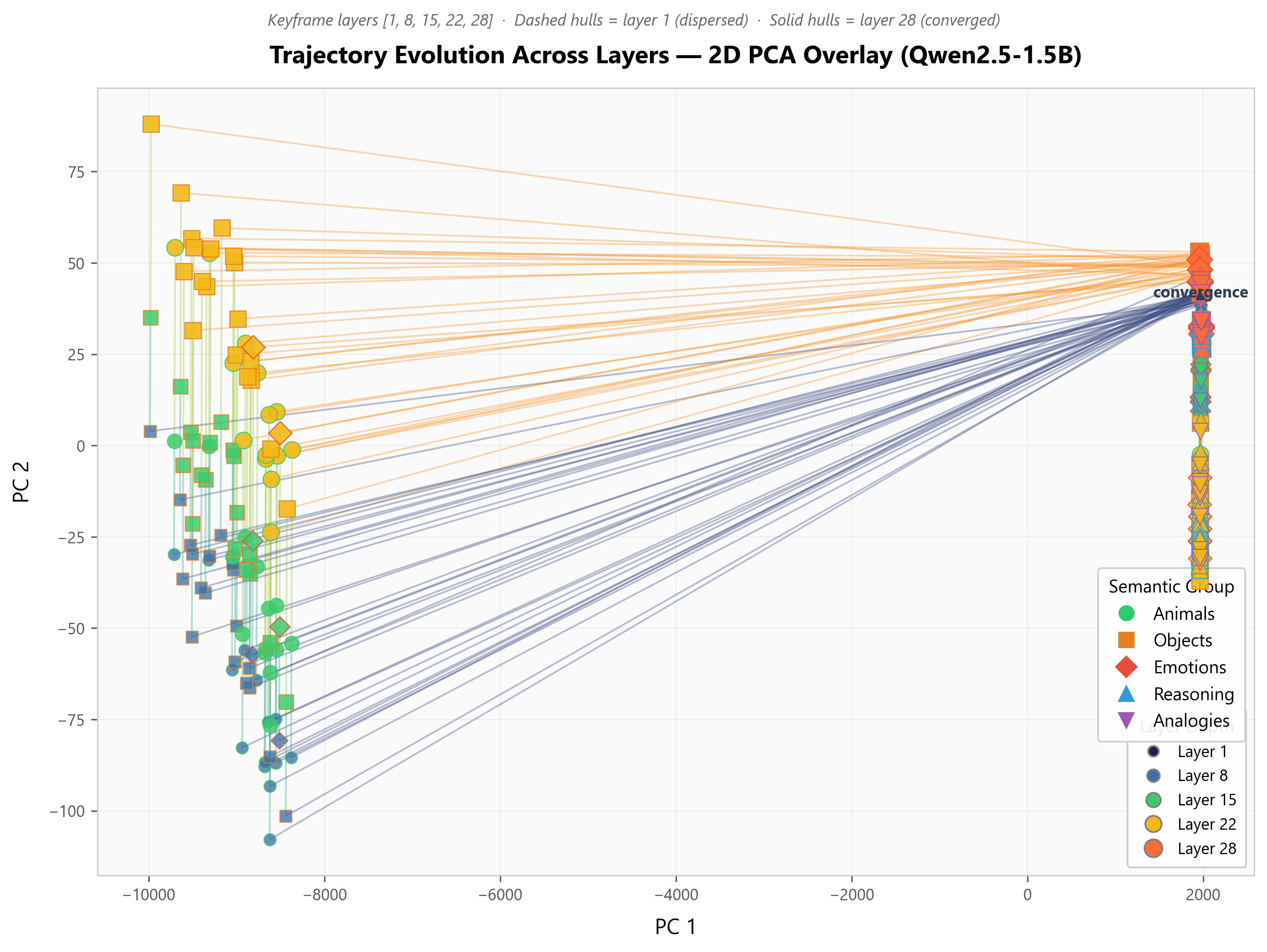}
\caption{
2D PCA overlay of trajectory keyframes across five selected layers of Qwen2.5-1.5B (28 layers total).
All five semantic groups are shown simultaneously; trajectory lines connect each prompt's representations across normalized keyframe depths.
The trajectory geometry converges consistently to tight semantic clusters in late layers, replicating the convergence structure observed in both GPT-2 and TinyLlama. This cross-architecture consistency demonstrates the robustness and generality of the trajectory-geometric framework: the intrinsic geometry of representation evolution is independent of model capacity and architectural details, suggesting it reflects fundamental properties of the learned semantic manifold.
}
\label{fig:app_keyframes_qwen25}
\end{figure}

\end{document}